\def\year{2019}\relax
\pdfoutput=1
\documentclass[letterpaper]{article} 
\usepackage{aaai19}  
\usepackage{times}  
\usepackage{helvet}  
\usepackage{courier}  
\usepackage{url}  
\usepackage{graphicx}  
\usepackage{amsmath}
\usepackage{amssymb}
\usepackage{bm}
\usepackage{color}
\usepackage{url}
\usepackage{subfig}

\begin{document}
%
\title{Using Monte Carlo Tree Search as a Demonstrator within Asynchronous Deep RL}
\author{Bilal Kartal\thanks{Equal contribution}, Pablo Hernandez-Leal$^*$ and Matthew E. Taylor\\
\texttt{\small\{bilal.kartal,pablo.hernandez,matthew.taylor\}@borealisai.com}\\
Borealis AI, Edmonton, Canada}
\maketitle 

\begin{abstract}

Deep reinforcement learning (DRL) has achieved great successes in recent years with the help of novel methods and higher compute power. However, there are still several challenges to be addressed such as convergence to locally optimal policies and long training times. In this paper, firstly, we augment Asynchronous Advantage Actor-Critic (A3C) method with a novel self-supervised auxiliary task, i.e. \emph{Terminal Prediction}, measuring temporal closeness to terminal states, namely A3C-TP.  Secondly, we propose a new framework where planning algorithms such as Monte Carlo tree search or other sources of (simulated) demonstrators can be integrated to asynchronous distributed DRL methods. Compared to vanilla A3C, our proposed methods both learn faster and converge to better policies on a two-player mini version of the Pommerman game.

\end{abstract}
 
\section{Introduction}

\noindent Deep reinforcement learning (DRL) combines reinforcement learning~\cite{sutton1998introduction} with deep learning~\cite{lecun2015deep}, enabling better scalability and generalization for challenging domains. DRL has been one of the most active areas of research in recent years with great successes such as mastering Atari games from raw images~\cite{mnih2015human}, AlphaGo Zero~\cite{silver2017mastering} for a Go playing agent skilled well beyond any human player (a combination of Monte Carlo tree search and DRL), and very recently, great success in DOTA 2~\cite{openfive}. For a more general and technical overview of DRL, please see the recent surveys~\cite{arulkumaran2017deep,li2017deep,hernandez2018multiagent}.

On the one had, one of the biggest challenges for reinforcement learning is the sample efficiency~\cite{yu2018towards}. However, once a DRL agent is trained, it can be deployed to act in real-time by only performing an inference through the trained model. On the other hand, planning methods such as Monte Carlo tree search (MCTS)~\cite{browne2012survey} do not have a training phase, but they perform simulation based rollouts assuming access to a simulator to find the best action to take.




There are several ways to get the best of both DRL and search methods. For example, AlphaGo Zero~\cite{silver2017mastering} and Expert Iteration~\cite{anthony2017thinking} concurrently proposed the idea of combining DRL and MCTS in an imitation learning framework where both components improve each other. These works combine search and neural networks \emph{sequentially}. First, search is used to generate an expert move dataset which is used to train a policy network~\cite{guo2014deep}. Second, this network is used to improve expert search quality~\cite{anthony2017thinking}, and this is repeated in a loop. However, expert move data collection by vanilla search algorithms can be slow in a sequential framework~\cite{guo2014deep}. 

In this paper, we show that it is also possible to blend search with distributed DRL methods such that search and neural network components are decoupled and can be executed \emph{simultaneously} in an \textit{on-policy} fashion. The planner (MCTS or other methods~\cite{lavalle2006planning}) can be used as a demonstrator to speed up learning for model-free RL methods. Distributed RL methods enable efficient exploration and yield faster learning results and several methods have been proposed~\cite{mnih2016asynchronous,jaderberg2016reinforcement,schulman2017proximal,espeholt2018impala}. 


In this paper, we consider Asynchronous Advantage Actor-Critic (A3C)~\cite{mnih2016asynchronous} as a baseline algorithm. We augment A3C with  \emph{auxiliary tasks}, i.e., additional tasks that the agent can learn without extra signals from the environment besides policy optimization, to improve the performance of DRL algorithms.

\begin{figure}
\centering
\includegraphics[scale=0.35]{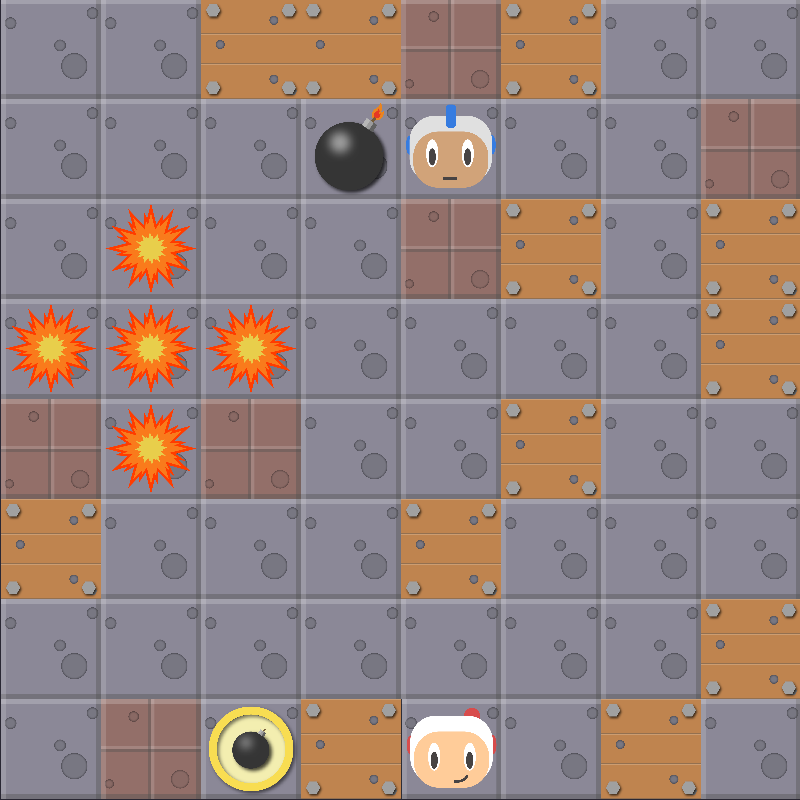}
\caption{An example of the $8 \times 8$ Pommerman board, randomly generated by the simulator. Agents' initial positions are randomly selected among four corners at each episode.}
\label{fig:pom8x8}
\end{figure}

\begin{itemize}
\item We propose a novel auxiliary task, namely \emph{Terminal Prediction}, based on temporal closeness to terminal states on top of the A3C algorithm. Our method, named A3C-TP, both learns faster and helps to converge to better policies when trained on a two-player mini version of the Pommerman game, depicted in Figure~\ref{fig:pom8x8}.

\item We propose a new framework based on diversifying some of the workers of A3C method with MCTS based planners (serving as demonstrators) by using the parallelized asynchronous training architecture to improve the training efficiency.
\end{itemize}

\section{Related Work}  

Our work lies at the intersection of the research areas of deep reinforcement learning methods, imitation learning, and Monte Carlo tree search based planning. In this section, we will mention some of the existing work in these fields.

\subsection{Deep Reinforcement Learning}

Reinforcement learning (RL) seeks to maximize the sum of discounted rewards an agent collects by interacting with an environment. RL approaches mainly fall under three categories: value based methods such as Q-learning~\cite{watkins1992q} or Deep-Q Network~\cite{mnih2015human}, policy based methods such as REINFORCE~\cite{williams1992simple}, and a combination of value and policy based techniques, i.e. actor-critic methods~\cite{konda2000actor}. Recently, there have been several distributed actor-critic based DRL algorithms~\cite{mnih2016asynchronous,jaderberg2016reinforcement,espeholt2018impala,gruslys2018reactor}.


A3C (Asynchronous Advantage Actor Critic)~\cite{mnih2016asynchronous} is an algorithm that employs a \emph{parallelized} asynchronous training scheme (using multiple CPU cores) for efficiency. It is an on-policy RL method that does not use an experience replay buffer. A3C allows multiple workers to simultaneously interact with the environment and compute gradients locally. All the workers pass their computed local gradients to a global neural network that performs the optimization and synchronizes with the workers asynchronously. There is also the A2C (Advantage Actor-Critic) method that combines all the gradients from all the workers to update the global neural network \emph{synchronously}. 

The UNREAL framework~\cite{jaderberg2016reinforcement} is built on top of A3C. In particular, UNREAL proposes unsupervised \emph{auxiliary tasks} (e.g., reward prediction) to speed up learning which require no additional feedback from the environment. In contrast to A3C, UNREAL uses an experience replay buffer that is sampled with more priority given to positively rewarded interactions to improve the critic network.




\subsection{Monte Carlo Tree Search} 

\begin{figure}[t!]
\centering
\includegraphics[width=0.21\linewidth]{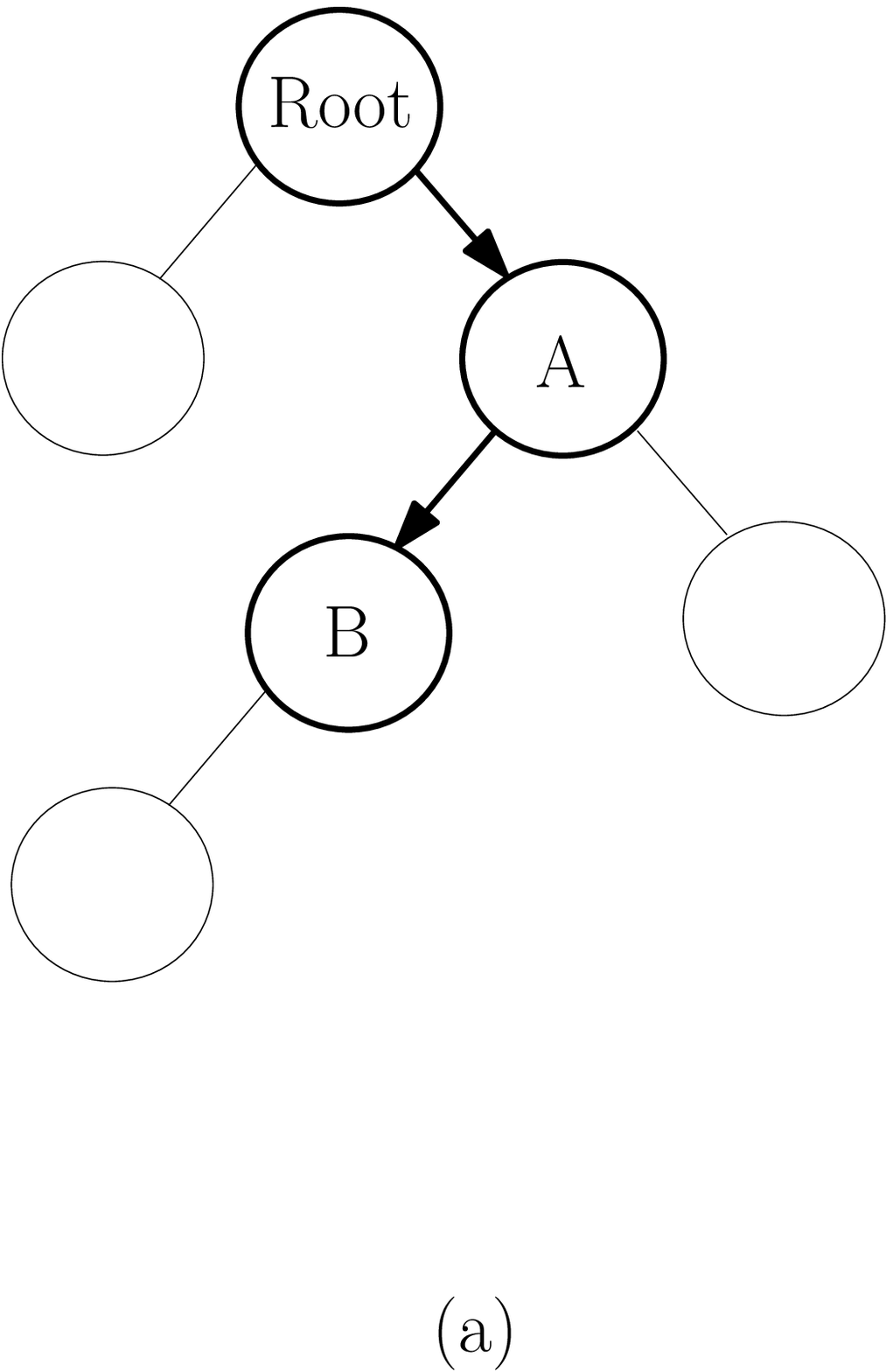}
\label{fig:selection}
\quad
\includegraphics[width=0.21\linewidth]{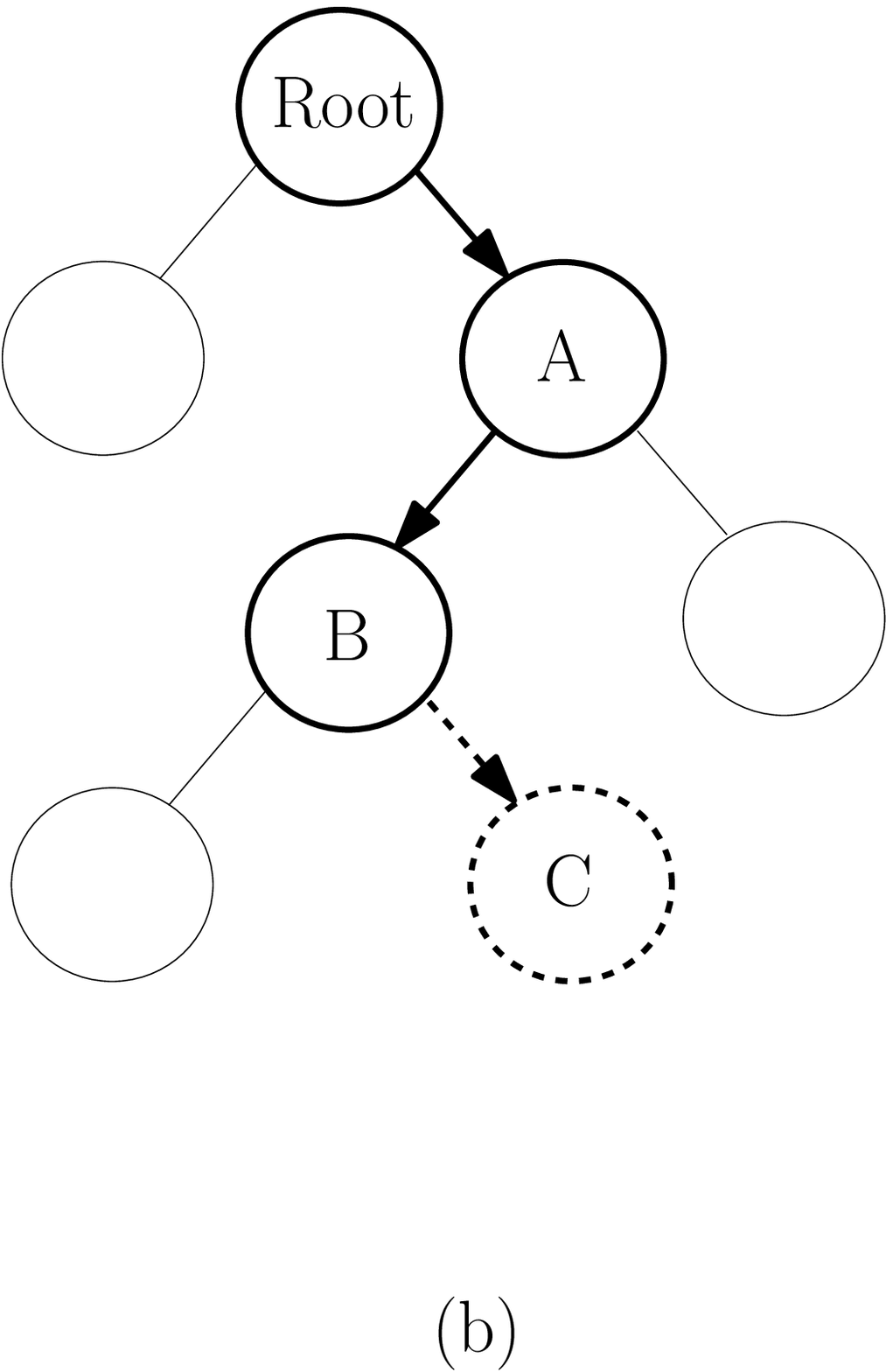}
\label{fig:expansion}
\quad
\includegraphics[width=0.21\linewidth]{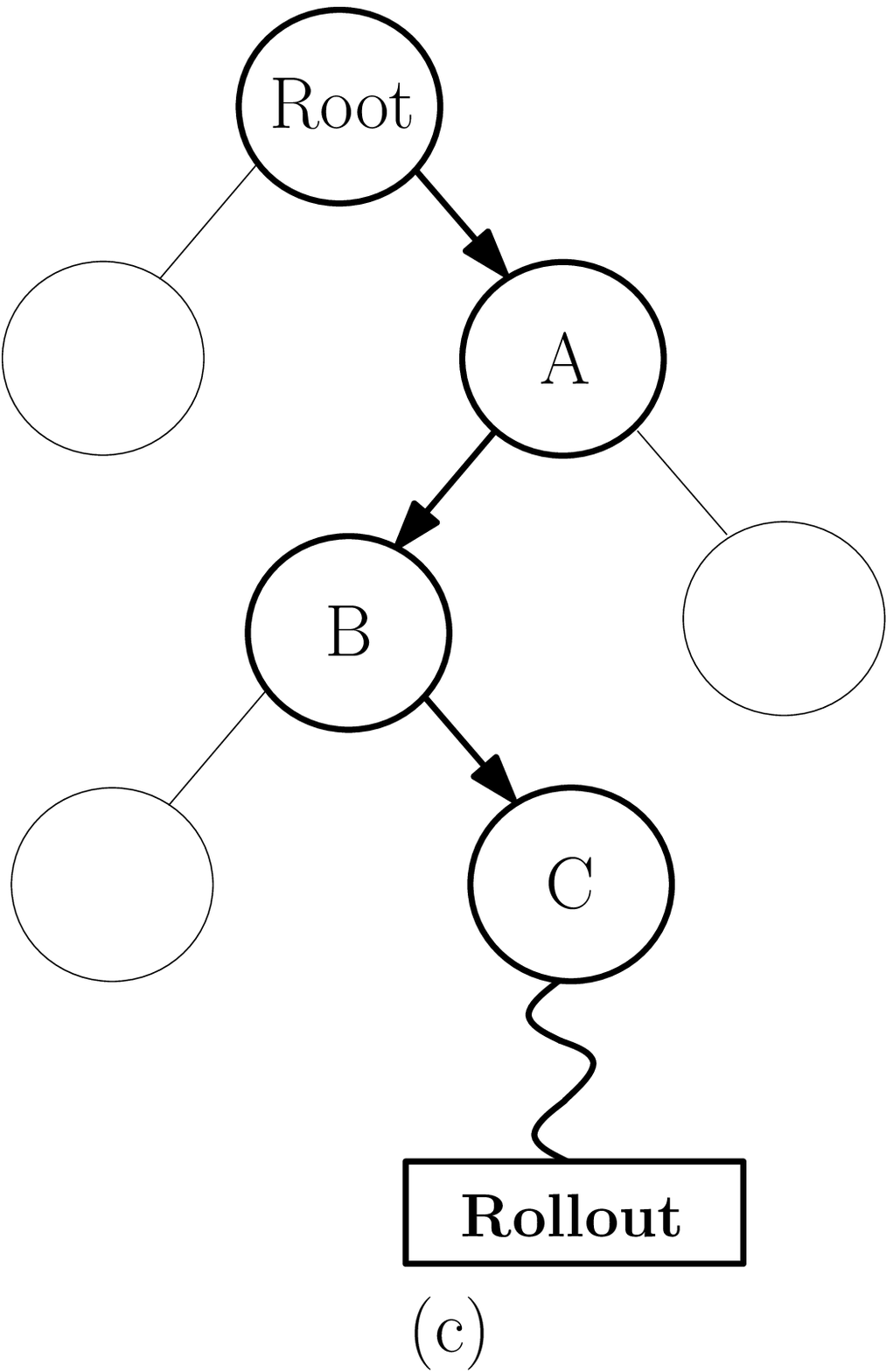}
\label{fig:rollout}
\quad
\includegraphics[width=0.21\linewidth]{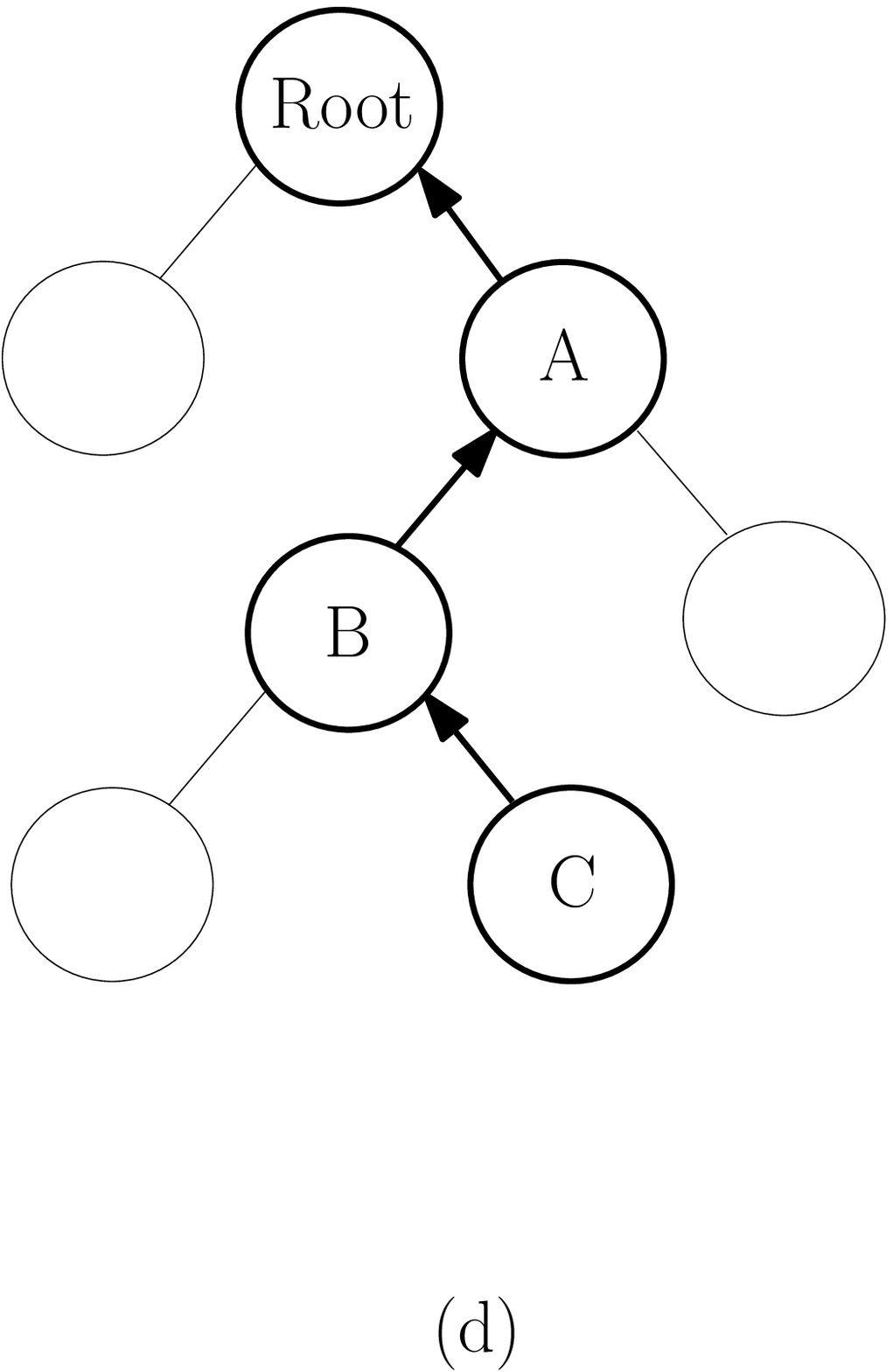}
\label{fig:backprop}
\quad
\caption{ Overview of Monte Carlo Tree Search: 
(a) \textit{Selection}: UCB is used recursively until a node with an unexplored action is selected. Assume that nodes A and B are selected. 
(b) \textit{Expansion}: Node C is added to the tree. 
(c) \textit{Random Rollout}: A sequence of random actions is taken from node C to complete the partial game. 
(d) \textit{Back-propagation}: After rollout terminates, the game is evaluated and the score is back-propagated from node C to the root.}
\label{fig:MCTS}
\end{figure}

Monte Carlo Tree Search (see Figure~\ref{fig:MCTS}) is a best first search algorithm that gained traction after its breakthrough performance in Go \cite{coulom2006efficient}. Other than for game playing agents, MCTS has been employed for a variety of domains such as robotics~\cite{kartal2016monte,zhang2017active} and Sokoban puzzle generation~\cite{kartal2016data}. A recent work~\cite{vodopivec2017monte} provided an excellent unification of MCTS and RL.

\subsection{Imitation Learning} 

Domains where rewards are delayed and sparse are difficult exploration RL problems and are particularly difficult when learning \textit{tabula rasa}. Imitation learning can be used to train agents much faster compared to learning from scratch. 

Approaches such as DAGGER~\cite{ross2011reduction} or its  extended version~\cite{sun2017deeply} formulate imitation learning as a supervised problem where the aim is to match the performance of the demonstrator. However, performance of agents using these methods is upper-bounded by the demonstrator performance.

Previously, Lagoudakis et al.~(\citeyear{lagoudakis2003reinforcement}) proposed a classification-based RL method using Monte-Carlo rollouts for each action to construct a training dataset to improve the policy iteratively. Other more recent works such as Expert Iteration~\cite{anthony2017thinking} extend imitation learning to the RL setting where the demonstrator is also continuously improved during training. There has been a growing body of work on imitation learning where human or simulated demonstrators' data is used to speed up policy learning in RL~\cite{hester2017deep,subramanian2016exploration,cruz2017pre,christiano2017deep,nair2018overcoming}. 

Hester et al.~(\citeyear{hester2017deep}) used demonstrator data by combining the supervised learning loss with the Q-learning loss within the DQN algorithm to pretrain and showed that their method achieves good results on Atari games by using a few minutes of game-play data. Cruz et al.~(\citeyear{cruz2017pre}) employed human demonstrators to pretrain their neural network in a supervised learning fashion to improve feature learning so that the RL method with the pretrained network can focus more on policy learning, which resulted in reducing training times for Atari games. Kim et al.~(\citeyear{kim2013learning}) proposed a learning from demonstration method where limited demonstrator data is used to impose constraints on the policy iteration phase and they theoretically prove bounds on the Bellman error.

In some domains, such as robotics, the tasks can be too difficult or time consuming for humans to provide full demonstrations. Instead, humans can provide more sparse \textit{feedback}~\cite{loftin2014strategy,christiano2017deep} on alternative agent trajectories that RL can use to speed up learning. Along this direction, Christiano et al.~(\citeyear{christiano2017deep}) proposed a method that constructs a reward function based on data containing human feedback with agent trajectories and showed that a small amount of non-expert human feedback suffices to learn complex agent behaviours. 

\subsection{Combining Search, DRL, and Imitation Learning}

We conclude our literature review with the description of the AlphaGo~\cite{silver2016mastering} and AlphaGo Zero~\cite{silver2017mastering} methods that combined methods from aforementioned research areas for a breakthrough success.

AlphaGo defeated the strongest human Go player in the world on a full-size board. It uses imitation learning by pretraining RL's policy network from human expert games with supervised learning~\cite{lecun2015deep}. Then, its policy and value networks keep improving by self-play games via DRL. And finally, an MCTS search skeleton is employed where a policy network narrows down move selection (i.e., effectively reducing the branching factor) and a value network helps with leaf evaluation (i.e., reducing the number of costly rollouts to estimate state-value of leaf nodes). AlphaGo Zero dominated AlphaGo even though it started to learn \textit{tabula rasa}. AlphaGo Zero still employed the skeleton of MCTS algorithm, but it employed the value network for leaf node evaluation without any rollouts.





\section{Preliminaries}  

In this section, we provide some formal background on reinforcement learning. We start with the standard reinforcement learning setting of an agent interacting in an environment over a discrete number of steps. At time $t$ the agent in state $s_t$ takes an action $a_t$ and receives a reward $r_t$. The discounted return is defined as $R_{t:\infty} = \sum_{t=1}^\infty \gamma^t r_t$. State-value function, $V^\pi(s)=\mathbb{E}[R_{t:\infty}|s_t=s,\pi]$, is the expected return from state $s$ following a policy $\pi(a|s)$.



The A3C method, as an actor-critic algorithm, has a policy network (actor) and a value network (critic) where the actor is parameterized by $\pi(a|s;\theta)$ and the critic is parameterized by $V(s; \theta_v)$, which are updated as follows:
$$\triangle\theta = \nabla_\theta \log \pi(a_t|s_t; \theta) A(s_t, a_t; \theta_v), $$
$$\triangle\theta_v = A(s_t, a_t; \theta_v) \nabla_{\theta_v} V(s_t)$$
where, 
$$A(s_t, a_t; \theta_v) = \sum_k^{n-1} \gamma^kr_{t+k} + \gamma^n V(s_{t+n}) - V (s_t).$$
The policy and the value function are updated after every $t_{max}$ actions or when a terminal state is reached. It is common to use one softmax output for the policy $\pi(a_t|s_t; \theta)$ head and one linear output for the value function $V (s_t; \theta_v)$ head, with all non-output layers shared.

\begin{figure*}
\centering
\includegraphics[width=\linewidth]{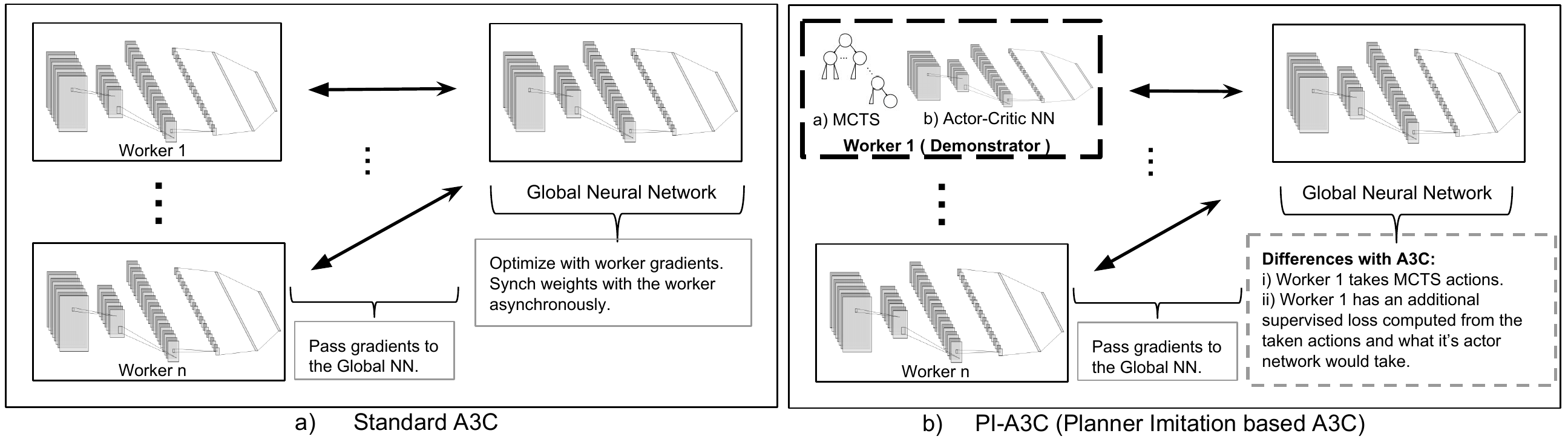}
\caption{\textbf{a)} In Asynchronous Advantage Actor-Critic (A3C) framework, each worker independently interacts with the environment and computes gradients. Then, each worker \emph{asynchronously} passes the gradients to the global neural network which updates parameters and synchronize with the respective worker. \textbf{b)} Our proposed framework, namely Planner Imitation based A3C (PI-A3C), is depicted. One worker is assigned as an MCTS based demonstrator taking MCTS actions while keeping track of what action it's actor network would take. The demonstrator worker has an additional auxiliary supervised loss different than the rest of the workers. PI-A3C enables the network to simultaneously optimize the policy and learn to imitate the MCTS.}
\label{fig:a3c_tp}
\end{figure*}

The loss function for A3C is composed mainly of two terms: policy loss (actor), $\mathcal{L}_{\pi}$, and value loss (critic), $\mathcal{L}_{v}$. An entropy loss for the policy, $H(\pi)$, is also commonly added which helps to improve exploration by discouraging premature convergence to suboptimal deterministic policies~\cite{mnih2016asynchronous}. Thus, the loss function is given by,

$$\mathcal{L}_{\text{A3C}} \approx  \mathcal{L}_{v} + \mathcal{L}_{\pi} - \mathbb{E}_{s \sim \pi} [H(\pi(s, \cdot, \theta)].  $$

As our work also augments A3C with auxiliary tasks, we also discuss the  
UNREAL algorithm, which is built on top of A3C. In particular, UNREAL proposes two auxiliary tasks: auxiliary control and auxiliary prediction, which both share the previous layers that the base agent uses to act. By using this jointly learned representation, the base agent learns to optimize extrinsic reward much faster and, in many cases, achieves better policies at the end of training.

The UNREAL framework optimizes a single combined loss function with respect to the joint parameters of the agent 
that combines the A3C loss, $\mathcal{L}_{A3C}$, together with an auxiliary control loss,  $\mathcal{L}_{PC}$, an auxiliary reward prediction loss, $\mathcal{L}_{RP}$, and a replayed value loss, $\mathcal{L}_{VR}$, as follows:
$$\mathcal{L}_{\text{UNREAL}}= \mathcal{L}_{A3C} + \lambda_{VR} \mathcal{L}_{VR} + \lambda_{PC} \mathcal{L}_{PC} + \lambda_{RP} \mathcal{L}_{RP}  $$
where $\lambda_{VR}$, $\lambda_{PC}$, and $\lambda_{RP}$ are weighting terms on the individual loss components. 

\section{Approach Overview}

In this section, we will present our two contributions. The first one, A3C-TP (Terminal Prediction), extends A3C with a novel auxiliary task of terminal state prediction which outperforms pure A3C. The second one is a framework that combines MCTS with A3C to accelerate learning. 

\subsection{Terminal State Prediction as an Auxiliary Task}

For domains with sparse and delayed rewards, RL methods require long training times when they are dependant on only external reward signals. UNREAL, by incorporating other loss terms (beyond external rewards), showed significant performance gains. UNREAL extends A3C with additional tasks that are optimized in an off-policy manner by keeping an experience replay buffer. In our work, we make minimal changes to A3C, we complement A3C's loss with an auxiliary task (i.e., terminal state prediction), but still require no experience replay buffer and keep our method fully on-policy. Thus, note that our approach is complementary to UNREAL.


We propose a new auxiliary task of terminal state prediction (i.e., for each observation under the current policy, the agent predicts a temporal closeness metric to a terminal state for the current episode). The neural network architecture of A3C-TP is identical to that of A3C, fully sharing parameters, except the additional terminal state prediction head outputs a value between $[0,1]$. 


We compute the loss term for the terminal state prediction head, $\mathcal{L}_{TP}$, by using mean squared error between the predicted probability of closeness to a terminal state of any given state (i.e., $y^p$) and the target values approximately computed from completed episodes (i.e., $y$) as follows: ${\mathcal{L}_{TP}= \frac{1}{N} \sum_{i=0}^{N}(y_{i} - y_{i}^p)^2}$ where $N$ represents the current episode length. We assume that the target for $i$th state can be approximated with $y_{i} = i/N $ implying $y_{N}=1$ for the actual terminal state and $y_{0}=0$ for the initial state for each episode, and intermediate values are linearly interpolated between $[0,1]$. 

Given $\mathcal{L}_{TP}$, we define loss for A3C-TP as follows: $$\mathcal{L}_{\text{A3C-TP}}= \mathcal{L}_{A3C} + \lambda_{TP} \mathcal{L}_{TP}$$ where $\lambda_{TP}$ is a weight term.

The hypothesis is that the terminal state prediction, as an auxiliary task, provides some grounding to the neural network during learning with a denser signal as the agent learns not only to maximize rewards but it can also predict approximately how close it is to the end of episode, when the reward signal will be received. This has been recently studied in the context of representation learning. For example, Schelhamer et al.~(\citeyear{shelhamer2016loss}) mentions that self-supervised auxiliary losses broaden the horizons of RL agents to learn from all experience (rewarded or not). One example is the reward which can be cast into a proxy task, which is expected to closely mirror the degree of policy improvement.

\subsection{Speeding up Training by Using MCTS as a Demonstrator}

As a second contribution we propose a framework that can use planners, or other sources of demonstrators, along with asynchronous DRL methods to accelerate learning. Even though our framework can be generalized to a variety of planners and distributed DRL methods, we showcase our contribution using MCTS and A3C. Our use of MCTS follows the approach in \cite{kocsis2006bandit} employing the UCB (Upper Confidence Bounds) technique to balance exploration versus exploitation during planning. The MCTS algorithm is described in Figure~\ref{fig:MCTS}. During rollouts, we simulate all agents as random agents as in default unbiased MCTS. We perform limited-depth rollouts to reduce action-selection time.

The motivation for combining MCTS and asynchronous DRL methods stems from the need to improve training time efficiency even if the planner or the world-model by itself is very slow. In this work, we assumed the demonstrator and actor-critic networks are decoupled, i.e. a vanilla UCT planner is used as a black-box that takes observation and returns an action without any access to actor-critic networks. Using vanilla MCTS in this case is conceptually similar to UCTtoClassification~\cite{guo2014deep}. However, it used 10K rollouts per action selection to construct an expert dataset to train neural network policy. In our domain (Pommerman) 10K rollouts would take hours due to slow simulator and long horizon~\cite{matiisen2018pommerman}. In light of these challenges, we aim to show how vanilla MCTS with a small number of rollouts ($\approx$ 100) can still be employed in an \textit{on-policy} fashion to improve training efficiency for actor-critic RL.

Within A3C's asynchronous distributed architecture, all the CPU workers perform agent-environment interaction with their neural network policy networks. In our new framework, namely PI-A3C (Planner Imitation with A3C), we assign one CPU worker to perform MCTS based planning for agent-environment interaction based on the agent's observations, while also keeping track of what its neural network would perform for those observations. In this fashion, we both learn to imitate the MCTS planner and to optimize the policy. The main motivation for PI-A3C framework is to increase the number of agent-environment interactions with positive rewards for hard-exploration RL problems to improve training efficiency.

The planner based worker still has its own neural network with actor and policy heads, but action selection is performed by the planner while its policy head is used for loss computation. The MCTS planner based worker augments its loss function with the supervised loss for the auxiliary task of \emph{Planner Imitation} with ${\mathcal{L}_{PI}= -\frac{1}{N} \sum_i^N a^i_{o} \log (\hat{a}^i_{o}) }$  being the supervised cross entropy loss between the one-hot encoded action planner used, $a^i_{o}$ and the action the actor (with policy head) would take in case there was no planner, $\hat{a}^i_{o}$ for an episode of length $N$. The demonstrator worker loss after addition of \emph{Planner Imitation}, is defined by $\mathcal{L}_{\text{PI-A3C}}= \mathcal{L}_{A3C} + \lambda_{PI} \mathcal{L}_{PI} $ where $\lambda_{PI}$ is a weight term.\footnote{Both Guo et al.~(\citeyear{guo2014deep}) and Anthony et al.~(\citeyear{anthony2017thinking}) used MCTS moves as a learning target, referred to as \textit{Chosen Action Target}. Our \textit{Planner Imitation} loss is similar except we employed cross-entropy loss in contrast to a KL divergence based one.}
In PC-A3C the rest of the workers (not demonstrators) are kept unchanged, still using the policy head for action selection with the unchanged loss function.
\emph{Planner Imitation} can be combined with either pure A3C through $\mathcal{L}_{\text{PI-A3C}}$ or our first contribution, A3C-TP by adding the $\lambda_{TP} \mathcal{L}_{TP}$ components.


\section{Experiments and Results}

\begin{figure*}[!h]
    \centering
    \subfloat[Learning against a \emph{Static} opponent.]{{\includegraphics[scale=0.32]{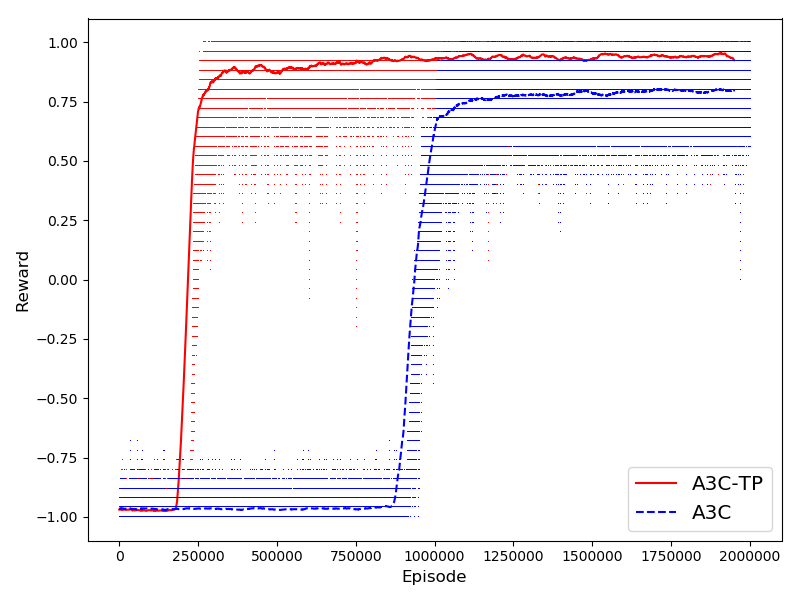} }}%
    \qquad\hspace{10mm}
    \subfloat[Learning against a \emph{Rule-based} opponent]{{\includegraphics[scale=0.32]{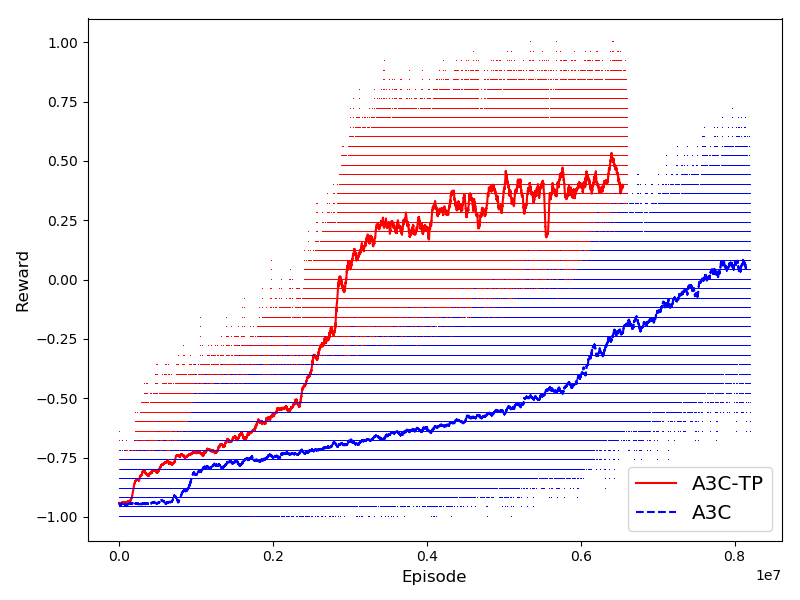} }}%
    \caption{Moving average over 50k games of the rewards (horizontal lines depict individual episodic rewards) is shown. Our method, A3C-TP, outperforms the standard A3C in terms of both learning faster and converging to a better policy in learning against both \emph{Static} and \emph{Rule-based} opponents. The training times was 6 hours for (a) and 3 days for (b).}%
    \label{fig:a3c_static_simple}%
\end{figure*}

This section describes the Pommerman two-player game used in the experiments. We then present the experimental setup and results against different opponents.

\subsection{Pommerman}

The Pommerman environment~\cite{resnick2018pommerman} is based off of the classic console game Bomberman. Our experiments use the simulator in a mode with two agents 
(see Figure~\ref{fig:pom8x8}). Each agent can execute one of 6 actions at every timestep: move in any of four directions, stay put, or place a bomb. Each cell on the board can be a passage, a rigid wall, or wood. The maps are generated randomly, albeit there is always a guaranteed path between any two agents. 
Whenever an agent places a bomb it explodes after 10 timesteps, producing flames that have a lifetime of 2 timesteps. Flames destroy wood and kill any agents within their blast radius. When wood is destroyed either a passage or a power-up is revealed. Power-ups can be of three types: increase the blast radius of bombs, increase the number of bombs the agent can place, or give the ability to kick bombs. A single game is finished when an agent dies or when reaching 800 timesteps.

Pommerman is a very challenging benchmark for RL methods. The first challenge is that of sparse and delayed rewards. The environment only provides a reward when the game ends (with a maximum episode length of 800), either 1 or -1 (when both agents die at the same timestep they both get -1). A second issue is the randomization over the environment since tile locations and agents' initial locations are randomized at the beginning of every game episode. The game board changes within each episode too, due to disappearance of wood, and appearance/disappearance of power-ups, flames, and bombs. The last complication is the multiagent component. The agent needs to best respond to any type of opponent, but agents' behaviours also change based on the collected power-ups, i.e., extra ammo, bomb blast radius, and bomb kick ability. For these reasons, we consider this game challenging for many standard RL algorithms and a local optimum is commonly learned, i.e., not placing bombs~\cite{resnick2018pommerman}.

Some other recent works also used Pommerman as a test-bed, for example, Zhou et al.~(\citeyear{zhou2018hybrid}) proposed a hybrid method combining rule-based heuristics with depth-limited search. Resnick et al.~(\citeyear{resnick2018backplay}) proposed a framework that uses a single demonstration to generate a training curriculum for sparse reward RL problems (assuming episodes can be started from arbitrary states); indeed, the same method is concurrently proposed for the game of Montezuma's Revenge~\cite{openaimr}.


\subsection{Setup}

We considered two types of opponents in our experiments:
\begin{itemize}
    \item \emph{Static} opponents: the opponent waits in the initial position and always executes the `stay put' action. 
    \item \emph{Rule-based} opponents: this is the baseline agent within the simulator. It collects power-ups and places bombs when it is near an opponent. It is skilled in avoiding blasts from bombs. It uses Dijkstra's algorithm on each time-step, resulting in longer training times.

\end{itemize}

\subsection{Results}

We present comparisons of our proposed methods based on the training performance in terms of converged policies and time-efficiency against \emph{Static} and \emph{Rule-based} opponents. All approaches were trained using 24 CPU cores. Unless otherwise noted, all approaches were trained for 3 days.

\subsubsection{Comparing A3C and A3C-TP}

The training results for A3C and our proposed method, A3C-TP, against a \emph{Static} opponent are shown in Figure~\ref{fig:a3c_static_simple}~(a). Our method both converges much faster, and to a better policy, compared to the standard A3C. The \emph{Static} opponent is the simplest possible opponent (ignoring suicidal opponents) for Pommerman as it provides a more stationary environment for RL methods. The trained agent needs to learn to successfully place a bomb near the enemy and stay out of death zone due to upcoming flames.

The other nice property of \emph{Static} opponents is that they do not commit suicide, and thus there are no false positives in observed positive rewards even though there are false negatives due to our agent's possible suicides during training. We consider false positive episodes when our agent gets a reward of +1 because the opponent commits suicide  (not due to our agent's combat skill), and false negatives episodes when our agent gets a reward of -1 due to its own suicide. False negative episodes are a major bottleneck for learning reasonable behaviours with pure-exploration within the RL formulation. Besides, false positive episodes also can reward agents for arbitrary passive survival policies such as camping or navigating on the board rather than engaging actively with opponents. These two cases render policy learning more challenging.

We also trained A3C and A3C-TP against the \emph{Rule-based} opponent. The results are presented in Figure~\ref{fig:a3c_static_simple}~(b), showing that our method learns faster, and it finds a better policy in terms of average rewards. Training against \emph{Rule-based} opponents takes much longer possibly due to episodes with false positive rewards. The \emph{Rule-based} agent's behaviour is stochastic, and based on the power-ups it collected, its behaviour further changes. For example, if it collected several ammo power-ups, it can place many bombs triggering chain explosions, i.e. some bombs explode earlier than their expected timestep due to being on a flame zone created by an exploded bomb.

\subsubsection{MCTS based Demonstrator}

\begin{figure*}[t]
    \centering
    \subfloat[A3C with and without a demonstrator.]{{\includegraphics[scale=0.32]{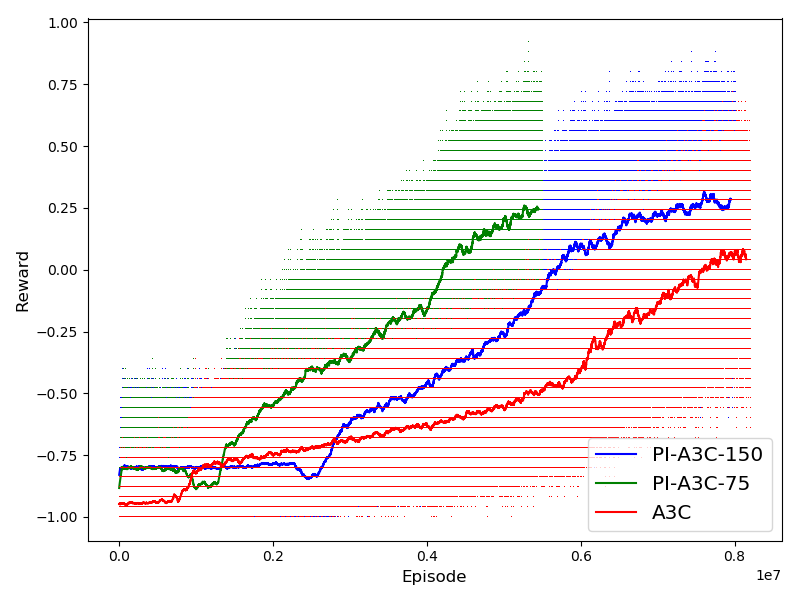} }}%
    \qquad\hspace{17mm}
    \subfloat[A3C-TP with and without a demonstrator.]{{\includegraphics[scale=0.32]{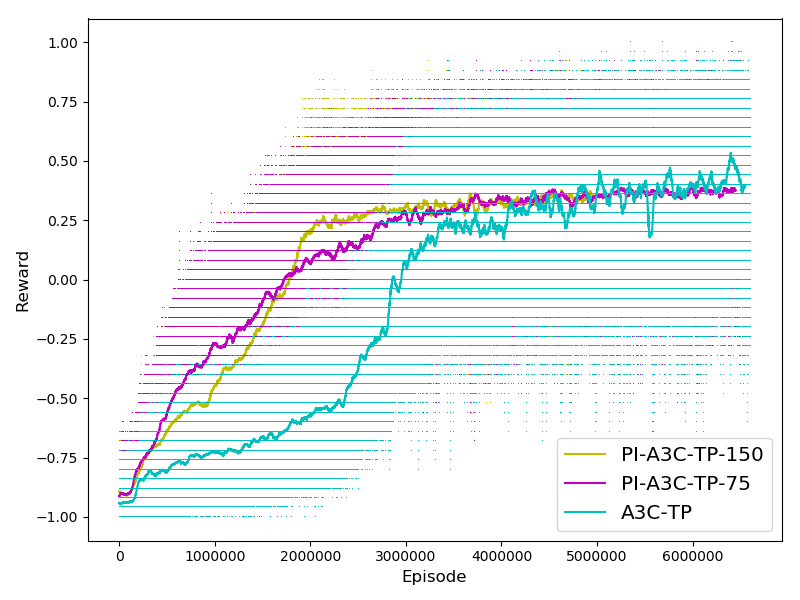} }}%
    \caption{Both figures were obtained by training against the \emph{Rule-based} opponent for 3 days. a) The ${\text{PI-A3C}}$ framework using MCTS demonstrator with 75 and 150 rollouts learns faster compared to the standard A3C. b)  The ${\text{PI-A3C-TP}}$ framework, where MCTS demonstrator is used for 75 and 150 rollouts, provides faster learning compared to our first contribution, A3C-TP. }%
    \label{fig:ablate_planner}%
\end{figure*}

To ablate the contribution of the demonstrator framework, we conducted two sets of experiments learning against \emph{Rule-based} opponents. Firstly, we compared the standard A3C method with PI-A3C. We present the results of this ablation in Figure~\ref{fig:ablate_planner}~(a), which shows that PI-A3C improves the training performance over our baseline, A3C. Secondly, we further extended A3C-TP method with the demonstrator framework, namely PI-A3C-TP. We present the results for this experiments in Figure~\ref{fig:ablate_planner}~(b), which shows the performance is further improved. Training curves for PI-A3C and PI-A3C-TP are obtained by using only the neural network policy network based workers, excluding MCTS based worker rewards.

We can vary the expertise level of MCTS as a demonstrator by changing the number of rollouts per action-selection. We experimented with 75 and 150 rollouts per move.

There is a trade-off in increasing the expert skill of the MCTS based demonstrator by increasing the planning time through rollouts --- the slower the planner, the fewer asynchronous updates will be made to the global neural network by the planner based worker, compared to the rest of the workers that use neural network policy head to act. From Figure~\ref{fig:ablate_planner}, we can see that the \emph{Planner Imitation} based methods with 75 rollouts, i.e., PI-A3C-75 and PI-A3C-TP-75, learn faster compared to the ones with 150 rollouts, as their demonstrator workers possibly make more updates to their global neural network, warming up worker actors better. However, at the end of training, the 150 rollout version converges to a similar or slightly better policy, compared to the 75 rollout version.

\section{Discussion}

\begin{figure}
\centering
\includegraphics[scale=0.32]{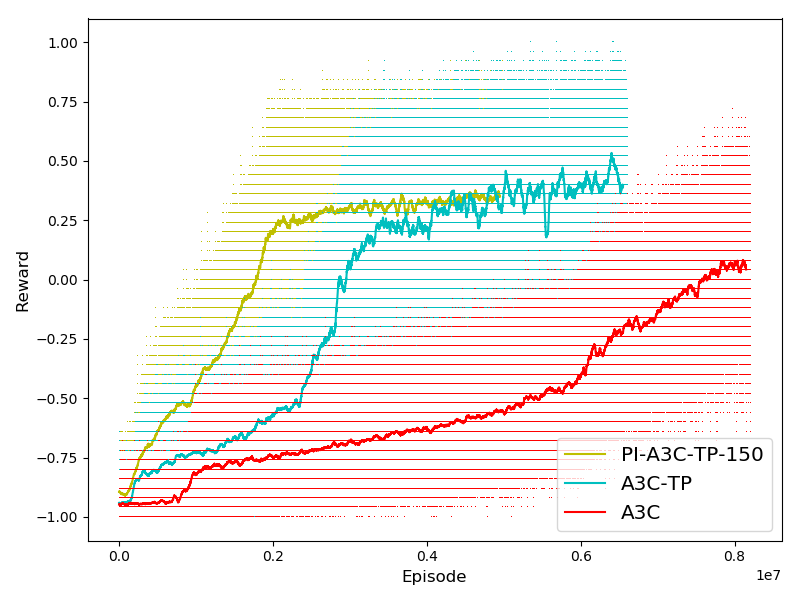}
\caption{Learning against a \emph{Rule-based} opponent. IM-A3C-TP, i.e. combining both of our contributions, learns faster compared to A3C and A3C-TP.}
\label{fig:combine_best_methods}
\end{figure}

To make the results clearer, we present A3C, A3C-TP, and PI-A3C-TP in Figure~\ref{fig:combine_best_methods}, showing that A3C-TP provides a significant speed-up in learning, in addition to converging to a better policy (compared to A3C). Moreover, combining A3C-TP with \emph{Planner Imitation}, namely PI-A3C-TP, converges with fewer episodes compared to A3C-TP.

In Pommerman, the main challenge for model-free RL is the high probability of suicide with a -1 reward while exploring, which occurs due to the delayed bomb explosions. However, the agent cannot succeed without learning how to stay safe after bomb placement. The methods and ideas proposed in this paper address this hard-exploration challenge. The first method that experimentally showed progress (as presented in this paper) is the framework that uses MCTS as a demonstrator, which yielded fewer training episodes with suicides as imitating MCTS helped for safer exploration. A second method is the \emph{ad-hoc} invocation of an MCTS based demonstrator within A3C that we will explain in detail. 

What we presented in this work clearly separates workers as either as planner/demonstrator-based or neural network-based. Another direction to test is the \emph{ad-hoc} usage of MCTS-based demonstrator within asynchronous DRL methods. Within an \emph{ad-hoc} combination, \emph{all workers} would use neural networks for action selection, except in ``relevant'' situations (based on some criteria) where they can pass the control to the simulated demonstrator. Indeed, the main motivation for our \emph{Terminal Prediction} auxiliary task was to use its prediction to decide \emph{when} to ask for a demonstrator action, thus, becoming the criteria to perform \emph{ad-hoc} demonstration. For example, if the \emph{Terminal Prediction} head predicts a value close to 1, this would signal the agent that episode is near a terminal state, i.e., a possible  suicide. Then, we would ask the demonstrator to take over in such cases to increase the number of episodes with positive rewards. However, the \emph{ad-hoc} invocation of demonstrators would be possible only when the \emph{Terminal Prediction} head outputs reliable values, i.e., after some warm-up training period. 

\subsubsection{Negative Results}

We attempted to adapt DQN and DRQN~\cite{hausknecht2015deep} to Pommerman, but they did not learn any reasonable behaviour, besides ``coward'' strategies where the agents learn to not place a bomb. 

We also adapted MCTS as a standalone planner without any neural network component for game playing. This played relatively well against \emph{Rule-based} opponents, but move decision time was well beyond 100ms limit\footnote{https://www.pommerman.com/competitions}.
\section{Future Work}

There are several directions to extend our work. In our current framework, we aimed to develop a framework where any simulated demonstrator as a black-box can be integrated to a distributed actor-critic RL method with the concept of auxiliary tasks in an on-policy fashion. Therefore, we employed vanilla MCTS as the demonstrator by keeping MCTS and A3C's actor and critic networks completely decoupled in contrast to AlphaGo Zero or Expert Iteration methods where MCTS actively uses neural networks. However, as these actor-critic networks are improved during training, MCTS could utilize them to speed up the search. Another direction is to experiment with multiple MCTS-based demonstrators compared to utilizing only one such worker.

All of our work presented in the paper is on-policy --- we maintain no experience replay buffer. This means that MCTS actions are used only once to update neural network and thrown away. In contrast, UNREAL uses a buffer and gives higher priority to samples with positive rewards. We could take a similar approach to save demonstrator's experiences to a buffer and sample based on the rewards.

In this work, to employ MCTS as a demonstrator, we assumed that we have access to a world model. However, there has been many works, e.g. the Dyna framework~\cite{sutton1991dyna}, that learn a world model while also optimizing agent policies. Another related recent work when an a priori world model is not accessible is Generative Adversarial Tree Search (GATS) method~\cite{azizzadenesheli2018sample}. GATS employs a Generative Adversarial Network (GAN) to learn a world model that MCTS can use to plan over.


\section{Conclusions} 

Deep reinforcement learning (DRL) has been progressing quite quickly in recent years. However, there are still several challenges to address, such as convergence to locally optimal policies and long training times. In this paper, we propose a new method, namely A3C-TP, extending the A3C method with a novel auxiliary task of \emph{Terminal prediction} that predicts temporal closeness to terminal states. We also propose a framework that combines MCTS as a simulated demonstrator within the A3C method, improving the learning performance within a two-player mini version of Pommerman game. Lastly, combining these two proposed methods obtained the best results. Although we showcase blending MCTS with A3C, this framework can be extended to other planners and distributed DRL methods.

\bibliographystyle{aaai}
\bibliography{ref}

\section{Appendix} 

\textbf{Neural Network Architecture:} For all methods described in the paper, we use a deep neural network with 4 convolutional layers, each of which has 32 filters and $3 \times 3$ kernels, with stride and padding of 1, followed with 1 dense layer with 128 hidden units, followed with 2-heads for actor and critic (where the actor output corresponds to probabilities of 6 actions, and the critic output corresponds to state-value estimate). For A3C-TP, the same neural network setup as A3C is used except there is an additional head for auxiliary \emph{terminal prediction}, which has a sigmoid activation function. After convolutional and dense layers, we used ELU  activation functions. Neural network architectures were not tuned.

\textbf{NN State Representation:} Similar to ~\cite{resnick2018pommerman}, we maintain 28 feature maps that are constructed from the agent observation. These channels maintain location of walls, wood, power-ups, agents, bombs, and flames. Agents have different properties such as bomb kick, bomb blast radius, and number of bombs. We maintain 3 feature maps for these abilities per agent, in total 12 is used to support up to 4 agents. We also maintain a feature map for the remaining lifetime of flames. All the feature channels can be readily extracted from agent observation except the opponents' properties and the flames' remaining lifetime, which can be tracked efficiently by comparing sequential observations for fully-observable scenarios.

\textbf{Hyperparameter Tuning:} We did not perform a through hyperparameter tuning due to long training times. We used a $\gamma=0.999$ for discount factor. For A3C, the default weight parameters are employed, i.e., $1$ for actor loss, $0.5$ for value loss, and $0.01$ for entropy loss. For new loss terms proposed in this paper, $\lambda_{TP}=1$ is used. For the \emph{Planner Imitation} task, $\lambda_{PI}=1$ is used for the MCTS worker, and $\lambda_{PI}=0$ for the rest of workers. We employed the Adam optimizer with a learning rate of $0.0001$. We found that for the Adam optimizer, $\epsilon = 1\times10^{-5}$ provides a more stable learning curve than its default value of $1\times10^{-8}$. We used a weight decay of $1\times10^{-5}$ within the Adam optimizer for L2 regularization.

\end{document}